\documentclass[conference]{IEEEtran}
\IEEEoverridecommandlockouts
\usepackage{cite}
\usepackage{amsmath,amssymb,amsfonts}
\usepackage{algorithmic}
\usepackage{graphicx}
\usepackage{textcomp}
\usepackage{xcolor}
\usepackage{multirow}
\usepackage[most]{tcolorbox}
\usepackage[utf8]{inputenc}
\usepackage{url}
\usepackage{xurl}

\def\BibTeX{{\rm B\kern-.05em{\sc i\kern-.025em b}\kern-.08em
    T\kern-.1667em\lower.7ex\hbox{E}\kern-.125emX}}
\begin{document}

\title{A Hybrid Search for Complex Table Question Answering in Securities Report
}

\author{
\IEEEauthorblockN{
  Daiki Shirafuji\IEEEauthorrefmark{1},
  Koji Tanaka\IEEEauthorrefmark{1},
  Tatsuhiko Saito
}
\IEEEauthorblockA{
Mitsubishi Electric Corporation,\\
Kamakura, Japan\\
\{Shirafuji.Daiki@ay., Tanaka.Koji@bc., Saito.Tatsuhiko@db.\}MitsubishiElectric.co.jp}
\thanks{\IEEEauthorrefmark{1}Authors are equally contributed.}
}

\maketitle
\IEEEpeerreviewmaketitle

\begin{abstract}
Recently, Large Language Models (LLMs) are gaining increased attention in the domain of Table Question Answering (TQA), particularly for extracting information from tables in documents. However, directly entering entire tables as long text into LLMs often leads to incorrect answers because most LLMs cannot inherently capture complex table structures. In this paper, we propose a cell extraction method for TQA without manual identification, even for complex table headers. Our approach estimates table headers by computing similarities between a given question and individual cells via a hybrid retrieval mechanism that integrates a language model and TF-IDF. We then select as the answer the cells at the intersection of the most relevant row and column. Furthermore, the language model is trained using contrastive learning on a small dataset of question-header pairs to enhance performance. We evaluated our approach in the TQA dataset from the U4 shared task at NTCIR-18. The experimental results show that our pipeline achieves an accuracy of 74.6\%, outperforming existing LLMs such as GPT-4o mini~(63.9\%). In the future, although we used traditional encoder models for retrieval in this study, we plan to incorporate more efficient text-search models to improve performance and narrow the gap with human evaluation results.
\end{abstract}

\IEEEpeerreviewmaketitle

\begin{IEEEkeywords}
Table Question Answering, Table Cell Search, Hybrid Search, Natural Language Processing, LLM
\end{IEEEkeywords}

\section{Introduction}

Tables in documents represent an important source of data in a wide range of domains,
such as finance~\cite{chen-etal-2021-finqa,koval-etal-2024-financial},
medical treatment~\cite{wang-sun-2022-promptehr},
and education~\cite{9856660}.
Techniques for extracting information from various types of tables
are essential in the field of natural language processing.

The rapid advancement of large language models (LLMs)
has led to numerous attempts to use LLMs
for table-related tasks~\cite{Lu_2024_tableLLM_survey,fang2024large}.
For instance,
some approaches directly input an entire table into LLMs for question answering~\cite{Li_TableGPT,chen_FewShotTableReasoners},
while others focus on cell-level information~\cite{glass-etal-2021-capturing}.

However, simply entering a whole table as a long text sequence
in LLM without considering
its structure can hinder the ability of the model
to capture the organization
of header rows and columns~\cite{glass-etal-2021-capturing}.
This issue becomes more notable with large tables
or complex layouts involving merged cells,
where LLMs alone often fail to generate correct answers.
Moreover, the computational cost of LLMs
increases exponentially
as the size of the input table, i.e., the sequence length, increases~\cite{sui-etal-2024-tap4llm,Tay_survey},
raising concerns about the cost efficiency and search precision of LLMs when dealing with large tables.

Kimura et al.~\cite{ntcir18-u4-overview} constructed
the Table Question Answering (TQA) dataset
for securities reports in accounting and finance,
including complex tabular data.
They aim to accurately extract the necessary values
from those tables with several LLMs.
However, lengthy finance tables
often have complex header structures,
so
results of directly entering a table into an LLM
would not be
sufficient~\cite{pang-etal-2024-uncovering,li-etal-2025-graphotter},
for example,
GPT-4o yields an accuracy of 64.75\%~\cite{ntcir18-u4-overview}.
This shows the need to identify
which cells in the table are required for the final answer.

In order to detect a correct cell for a question,
several works~\cite{Fang2012,Nagy2016,jauhar-etal-2016-tables,jin-etal-2023-enhancing,sun2016table}
try to identify header rows or columns, as they provide essential clues for locating relevant data.
However, existing methods rely on manually annotated table headers
or target the datasets with simple table formats
where headers can be identified through rule-based approaches.
They do not handle table data in HTML formats
that involve complexities,
such as merged cells and nested structures.

In this paper,
we propose a cell extraction method for the TQA dataset~\cite{ntcir18-u4-overview}
that does not assume the prior identification of the table headers.
\footnote{
Our approach is developed specifically for the Unifying, Understanding, and Utilizing Unstructured Data in Financial Reports~(U4) shared task in the NTCIR-18 conference~\cite{ntcir18-u4-overview}.
The experimental results presented here are taken from the official evaluation of that shared task.
}
Our approach estimates table headers by computing similarities
between a given question and individual cells,
using a hybrid retrieval method that combines a language model and TF-IDF.
The cells at the intersection of the most relevant row and column are then selected as the answer.

\begin{figure*}[t]
  \begin{center}
    \includegraphics[width=2\columnwidth]{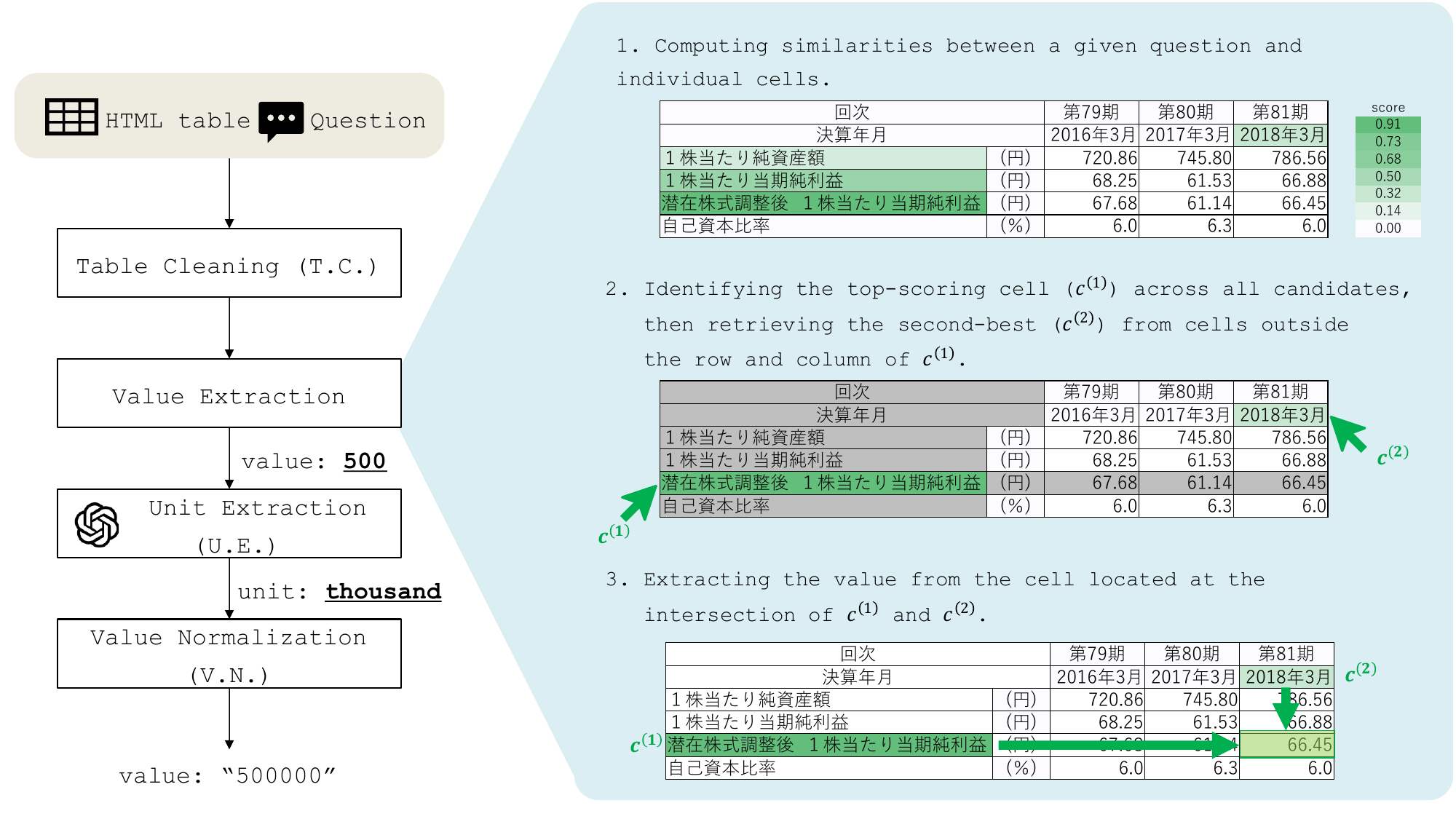}
    \caption{Overview of the proposed pipeline with examples of tabular data from the Japanese financial statements.}
    \label{fig:method_overview}
  \end{center}
\end{figure*}

An overview of our pipeline for the TQA task is shown in Fig.~\ref{fig:method_overview},
and consists of the following four steps:
(1) Cleaning tables (Section~\ref{sec3A});
(2) Identifying the cell corresponding to the question and retrieving its value using our proposed method (Section~\ref{sec3B});
(3) Locating the associated unit for the value from the given document (Section~\ref{sec3C});
and
(4) Answering a given question by normalizing the value with the retrieved unit (Section~\ref{sec3D}).

The experimental results on the TQA dataset demonstrate that
our proposed method outperforms the direct application of GPT-4o mini to the TQA dataset,
even with fine-tuning language models using 1,000 pairs of a question and a corresponding table
via contrastive learning.
This higher accuracy results from
resolving header positions and other intricate table layouts, even when they are complex.

Our main contributions are as follows.
\begin{itemize}
  \item Proposing a TQA pipeline that does not require prior identification of table headers.
  \item Demonstrating that our pipeline outperforms GPT-4o mini in the TQA dataset in the financial domain.
  \item Measuring how a human annotator without financial expertise performs on the TQA dataset in the financial domain using the TQA dataset, and comparing the resulting scores with those of LM-based methods.
\end{itemize}

\section{Related Work}

\subsection{Table Data Extraction}

A critical first step in extracting the required information
from the table data is the automatic identification of headers~\cite{Fang2012}.
Its headers typically refer to the rows or columns that indicate
its categories
and serve as clues for recognizing
where essential information is located in a table.


Fang et al.\cite{Fang2012} proposed a header detection method
that trains a model using features, such as string formatting and cell content patterns,
extracted from a large collection of tables available online.
The model classifies each row and column as headers or data.

Nagy et al.\cite{Nagy2016} focused on
the distribution of information within a table,
proposing a method to determine the boundary between headers and numerical values,
while also analyzing the hierarchical structure of tables.

The retrieval of relevant cells from tables for QA tasks
is also widely tackled~\cite{jauhar-etal-2016-tables,jin-etal-2023-enhancing}.
Sun et al.~\cite{sun2016table} proposed an approach
that, given a question (e.x., ``What is the population of Tokyo?'')
identifies the relevant row and column header
containing ``Tokyo'' and ``population,''
and regards the cell where these intersect as the answer.
However, this approach assumes that
terms appearing in the headers and the questions
are exactly the same
and that the headers within the table are already identified.

In recent years,
dense retrieval has been frequently introduced to improve search engines
for tabular data~\cite{yin-etal-2020-tabert,herzig-etal-2020-tapas,jin-etal-2023-enhancing}.
Jin et al.~\cite{jin-etal-2023-enhancing} trained LM representations
designed to match query keywords with the column structures of tables,
enabling efficient cell extraction from large-scale table corpora.
Utilizing dense retrieval,
they improved
recall scores
of cell extraction by capturing information
that traditional keyword-based retrieval methods would not acquire,
thereby overcoming the previous issue
that required consistent terminology across questions and tables, to some extent.


Most existing studies still rely on the assumption
that table headers are already identified
or that the terms appearing in the headers and
the questions match exactly.
Therefore, our proposed pipeline specifically addresses this limitation.

\subsection{LLM and Table Data}
Due to the influence of LLMs,
represented by GPT-4~\cite{openai2024gpt4technicalreport},
these models have increasingly been used for table data analysis~\cite{Lu2024TableSurvey}.
Since LLMs are pre-trained on massive text corpora  
that include knowledge derived from web tables,
the expectation that they can readily parse table structures
and accurately generate answers
has risen.


Recent works have attempted to address table QA
by directly prompting LLMs
or applying few-shot learning~\cite{Li_TableGPT,fang2024large,Lu_2024_tableLLM_survey,chen_FewShotTableReasoners}.
Wang et al.~\cite{wang2024chain} proposed a method
that applies Chain-of-Thought reasoning to tables~\cite{ChainOfThought}.
This approach makes an LLM output and update
intermediately generated tables during reasoning.
This work also reported that models produce incorrect reasoning steps
when complex reasoning, aggregation, or logical inference are required.


Wang et al.~\cite{wang2024chain} argue that
LLMs are increasingly applied to table data
and can locate the relevant cells for simple questions.
Opposite to this argument,
Kimura et al.\cite{ntcir18-u4-overview}
proposed a QA task for complex tables including in securities reports
and performed a comprehensive evaluation experiment with
representative LLMs,
such as GPT-4o~\cite{hurst2024gpt4o}, Gemini~\cite{team2023gemini}, and Claude 3~\cite{Claude3}.
They demonstrated that even for simple questions,
if the required table structure itself is complex,
the answer accuracy of these models remains insufficient.

These findings indicate that
numerous issues still need to be addressed in the use of LLMs for analyzing tabular data.

\section{Proposed Pipeline}
In this study,
we propose a pipeline
that targets securities reports that contain complex tables
and accurately extracts the necessary information from those tables.

An overview of our proposed pipeline is shown in Fig.~\ref{fig:method_overview}.
The process of our approach includes the following four steps:
(1) Table Cleaning (T.C.): Removing non-tabular information from the data,
(2) Value Extraction: Extracting the numerical data corresponding to the question from the table,
(3) Unit Extraction (U.E.): Retrieving the unit of the numerical data from the table,
and
(4) Value Normalization (V.N.): Outputting the answer using the numerical data and its unit information.

\subsection{Table Cleaning} \label{sec3A}
Tabular data generally contains not only the HTML structure of tables but also decorative information of tables and text that is actually non-tabular data.
Since leaving these non-tabular elements
can negatively impact cell extraction,
we clean data according to the following criteria.

First, we eliminate any text document outside the table, as well as annotation information unrelated to the table itself.
We secondly delete all HTML attributes
except ``colspan'' and ``rowspan'' --
e.g., ``style,'' ``class,'' and ``id.''
Preliminary experiments
with GPT-4o mini~\footnote{\url{https://openai.com/ja-JP/index/gpt-4o-mini-advancing-cost-efficient-intelligence/}.},
described in Appendix~\ref{app1},
showed that the presence of these decorative attributes could lead to misunderstanding
in retrieval; therefore, we apply this preprocessing step.
Finally, any additional HTML tags appearing inside
a ``td'' tag are removed,
converting the content into plain text.
In some cases, nested tables may appear within a single cell.
Although it would be ideal to treat such nested structures as separate tables,
we handle these cases by merging them into
a single cell to avoid the complexity of implementation
in these experiments.

\subsection{Value Extraction} \label{sec3B}
Following data cleaning,
we take the cleaned table data and the corresponding question text as input
to estimate the cell value that serves as the answer.

Our preliminary experiment with GPT-4o mini, described in Appendix~\ref{app1},
indicates that simply using a LLM for this step can be problematic:
when a table is large or highly structured,
the LLM may not grasp the structure of the table, resulting in lower accuracy of the responses.

To address this issue,
instead of directly identifying the answer cell by LLMs,
our approach estimates which row cell and column cell
are the most relevant and uses their intersection as the answer.

\subsubsection{Hybrid Search for Cell Scoring}
First, we treat the question and the content of each cell (i.e., each string) as a document $d$,
then perform the document retrieval.
We apply a hybrid method
combining TF-IDF with LM-based vector retrieval,
assigning a score $s_H(q,d)$
to each cell.
Specifically,
$s_H(q,d)$ is computed by taking a weighted average of the TF-IDF score and the vector-retrieval score
of the language model.
$s_H(q,d)$ is computed as follows:

$$
s_H(q, d) = (1 - \alpha) s_v(q, d) + \alpha s_t(q, d),
$$
where $q$ is a question, $d$ indicates a document,
and $\alpha$ is a hyperparameter to control the ratio of the scores of TF-IDF ($s_t(q,d)$) and vector-retrieval ($s_v(q,d)$).
To calculate $s_t(q,d)$,
we tokenize sentences with Mecab
\footnote{\url{https://taku910.github.io/mecab/#parse}.},
construct TF-IDF vectors using the scikit-learn library
\footnote{\url{https://scikit-learn.org/stable/modules/generated/sklearn.feature_extraction.text.TfidfVectorizer.html}.},
then calculate the cosine similarities $s_t(q,d)$ between vectors of $q$ and $d$.

For the vector-retrieval,
we vectorize sentences by language models with the SentenceTransformers library
\footnote{\url{https://www.sbert.net/}.},
and calculate the cosine similarity between the vectors of $q$ and $d$ as $s_t(q,d)$.

To determine the hyperparameter $\alpha$, we first evaluated the values from $0$ to $1$ in increments of $0.1$,
using accuracies on the validation dataset.
From these initial results, $\alpha=0.2$ achieved the highest validation accuracy.
We then performed a finer search between $0.1$ and $0.3$ at intervals $0.01$, 
ultimately selecting $\alpha=0.21$,
which yielded the best validation performance.

Thus, our pipeline provides a balanced assessment of both lexical~(TF-IDF)
and semantic~(vector retrieval).

\subsubsection{Identifying Corresponding Cells}

From the results of the hybrid search,
we select the two highest scoring cells,
labeled $c^{(1)}$ and $c^{(2)}$,
ensuring that they are not located in the same row and column.
Subsequently, the cell positioned more towards
the lower-right corner is assigned as the row-indicating cell while the other is designated as the column-indicating cell.

The intersection of the row and column is then deemed the answer to the question.
We expect this approach to robustly extract the correct cell
even if the header positions in tables are unclear,
or if there are merged cells or multi-level headers.

\subsubsection{Language Model Training}
To improve search accuracy, the language model for vector retrieval is better kept continually trained in the target domain.

Sentence-BERT~\cite{reimers-gurevych-2019-sentence} is widely used
to capture sentence meaning and transfer the model to the target domain.
Sentence-BERT employs contrastive learning by treating
the corresponding text pairs as positive examples,
with all other pairs serving as negatives.

\subsection{Unit Extraction} \label{sec3C}
The numerical values in the cells of the securities reports usually omit the associated unit,
such as ``thousands of yen,'' ``hundred shares,'' or ``\%.''
Even if a company's revenue is written simply as ``530,''
the context or other cells might suggest that it refers to ``thousands of yen,''or 
``hundreds of millions of yen.''

In this study, we use a prompt-based extraction approach with GPT-4o mini
to determine the units necessary for an accurate comparison and calculation of cell values.
Specifically, we input both the question text and the cleaned table,
prompting the model to output unit expressions inferred from all cells.

Our prompt is described in Appendix~B.

\subsection{Value Normalization} \label{sec3D}
Finally, we normalize the extracted numeric information together with the identified units
to convert it into an actual numerical representation.
In securities reports, a value corresponding to, for example, ``530,000''
 is often written simply as ``530.''

Therefore, we
(1) apply unit information
(e.g. ``thousands of yen'' $\to$ multiply by 1,000)
to adjust the scale 
and
(2) use a rule-based procedure to incorporate information.
For example,
when the unit is determined to be ``thousands of yen,''
a cell value ``530'' is corrected to ``530000'' yen.
This allows us to provide
an answer consistent with the final numeric format
required by the TQA dataset.

\section{Experiments}

\subsection{Evaluation Dataset}
We evaluate our proposed approach with the TQA dataset
provided by the shared task: U4~(Unifying, Understanding, and Utilizing Unstructured Data in Financial Reports)
in NTCIR-18~\cite{ntcir18-u4-overview}.

The dataset is divided into
train (10,300 samples),
validation (1,441 samples),
and test data (2,898 samples).
The TQA dataset provides
document information~(including tabular data),
a corresponding question,
and an answer~(the required cell ID and the final numerical output).
The purpose of this shared task is
to identify the correct cell number
and to result in the numerical answer.

Out of 10,300 samples in the original training dataset,
900 of them are allocated to train the language model,
and 100 were designated as validation data during training.
The original validation data is utilized for hyperparameter tuning in hybrid retrieval,
as described in Section~\ref{sec3B}.

\subsection{Evaluation Metrics}
Following Kimura et al.~\cite{ntcir18-u4-overview},
in order to measure how many cell numbers and answers are correct,
we adopt the accuracy for the extraction of value and cell id in an HTML table.

For both evaluation metrics,
correctness is determined based on an exact match to the reference.
Any output that differs from the correct answer even by a single character is regarded as incorrect.
For example, if the correct answer is “1,000,000” and the model outputs ``1 million yen,'' the output is considered incorrect.
In this evaluation, only the numeric component of the normalized value is treated as the correct answer;
supplementary expressions such as ``million yen'' are excluded from the evaluation.

\subsection{Target Pre-trained LMs}
In our experiments, we adopt two LMs:
Japanese Sentence-BERT~\cite{reimers-gurevych-2019-sentence}
\footnote{\url{https://huggingface.co/sonoisa/sentence-bert-base-ja-mean-tokens-v2}.}
as the baseline of language models trained on texts in the general domain,
and
a LUKE-based model~(UBKE-LUKE)~\cite{yamada-etal-2020-luke}~\footnote{\url{https://huggingface.co/uzabase/UBKE-LUKE}.},
which is an encoder model specialized for corporate entities that is trained on data from the economic information domain, especially securities reports.

\subsection{Tokenizer for TF-IDF Search Scoring}
In the proposed method, we compute the search score using TF-IDF.
For tokenization during the TF-IDF application, we utilize MeCab~\cite{Kudo2005MeCabY}
with the ipadic dictionary.

\begin{figure*}[t]
  \begin{center}
    \includegraphics[width=2\columnwidth]{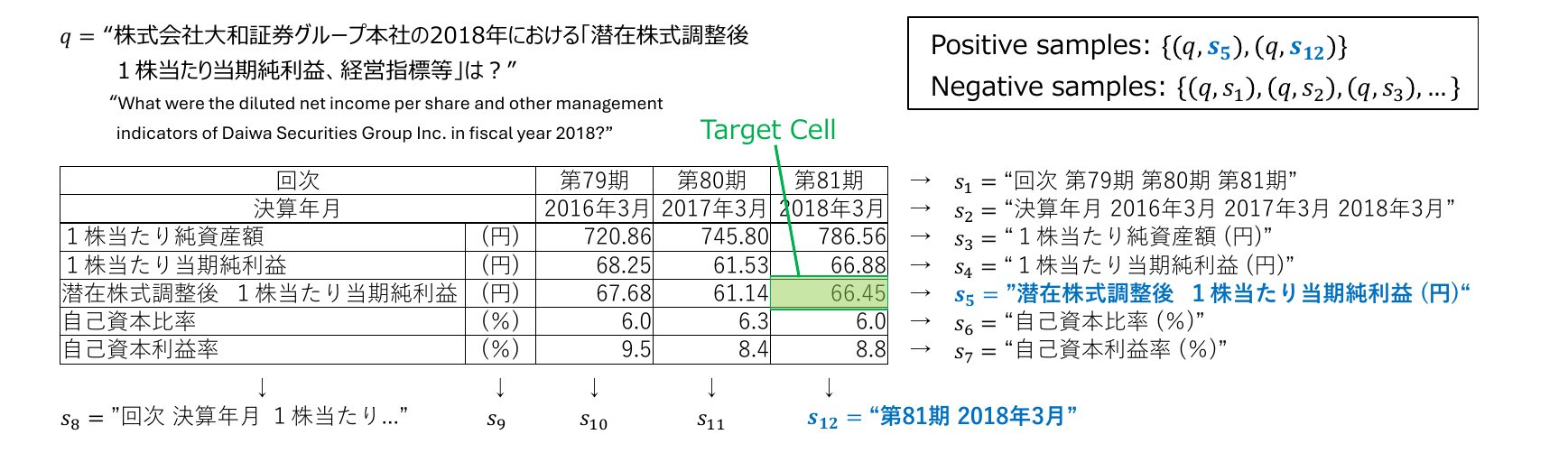}
    \caption{Overview of our method for constructing a training dataset for training Language Model based on the TQA dataset.}
    \label{fig:training_data}
  \end{center}
\end{figure*}

\subsection{Dataset for LM Training} \label{chap3-5}
The TQA dataset,
constructed by Kimura et al.~\cite{ntcir18-u4-overview}, 
only provides pairs of a question and a correct cell,
which complicates the direct construction of these pairs.

We build the pseudo-pairs of a question and a text obtained by concatenating all the cell values in each row and column.
The overview of data creation is shown in Figure~\ref{fig:training_data}.
If a row or column header contains the correct cell,
the pair is treated as a positive example;
otherwise, it is negative.

Numeric or symbol-only cells are excluded from the training,
focusing on cells containing Japanese or English words.

We create a pair of sentences from a part of the TQA train dataset~~\cite{ntcir18-u4-overview},
where we target 1,000 questions.
Our dataset finally contains 31,325 pairs;
2,000 pairs are positive, and the others are negative.
90\% of them are for training, and the others are for the validation dataset.

By learning which rows and columns
within the entire table are most relevant to the question,
the model can achieve highly accurate retrieval even without explicitly defined header rows or columns.

\subsection{Experimental Setup for LM Training}
We fine-tune Sentence-BERT on a pseudo dataset described in Section~\ref{chap3-5}
with the following hyperparameters in the experiments:

\begin{itemize}
  \item Batch size: 16,
  \item Learning rate: 2e-5,
  \item Scheduler for learning rate: linear,
  \item Optimizer: Adam~\cite{kingma2017},
  \item Warmup ratio: 0.1,
  \item Number of weight decays: 0.01.
\end{itemize}

We additionally apply early stopping
to avoid overfitting
when the model's performance on the validation dataset fails to improve.
We set the patience parameter for early stopping to 500 steps,
which means that if no improvement is observed for 500 consecutive steps,
the training process will be terminated.

\begin{figure*}[t]
  \begin{center}
    \includegraphics[width=1.5\columnwidth]{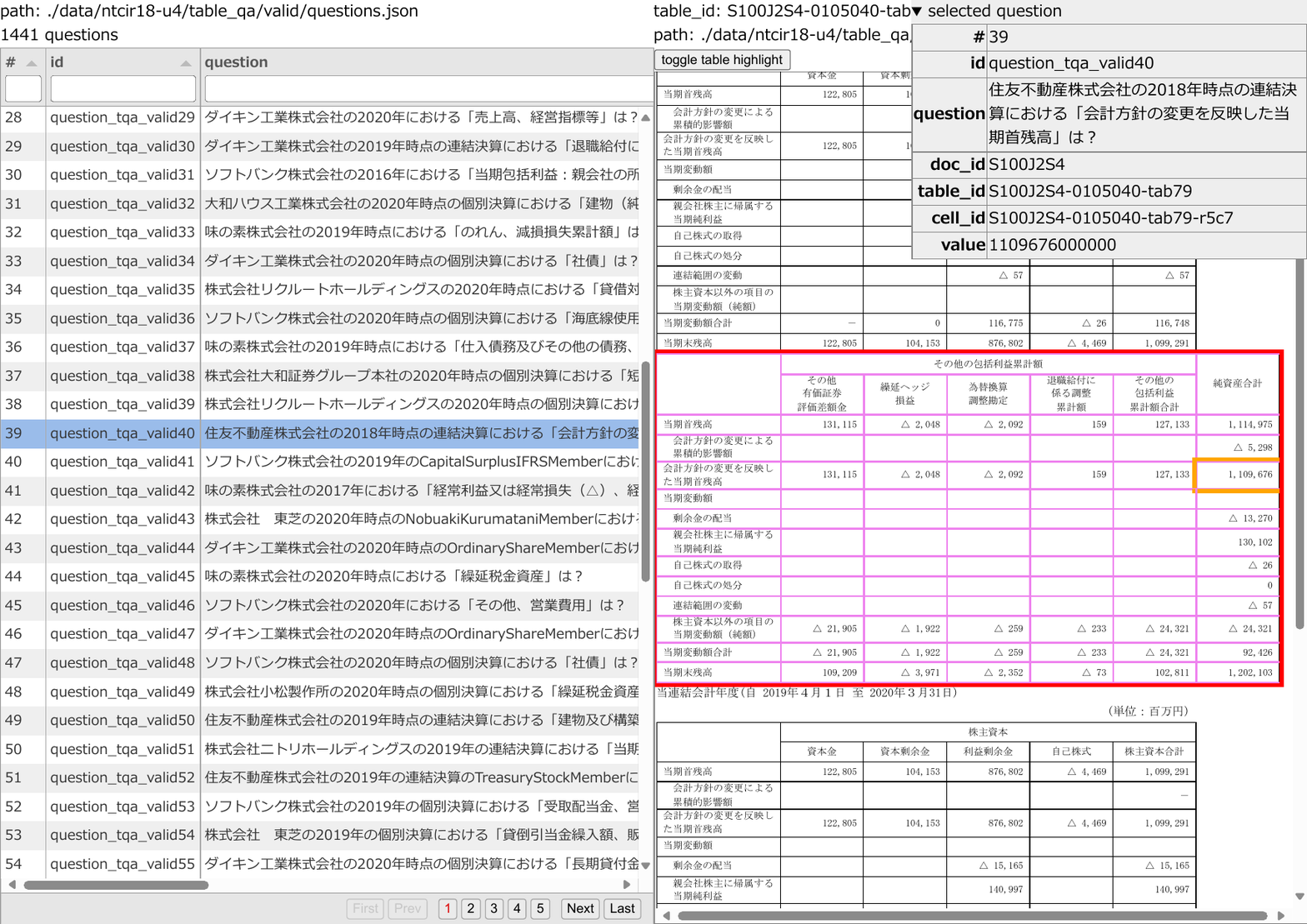}
    \caption{An example of our GUI for the human evaluation process.}
    \label{fig:gui}
  \end{center}
\end{figure*}

\subsection{Human Evaluation Process} \label{app3}

We perform a human evaluation of the test dataset in the TQA.
The annotator is a native Japanese speaker without any prior knowledge in the economic domain.

To facilitate this annotation process,
we develop a graphical user interface (GUI)
that allows the annotator to identify the cell IDs in the table corresponding to a given question.
An example of this interface is shown in Figure~\ref{fig:gui}.

When a question is selected from the list on the left side of the GUI,
the corresponding table is displayed on the right side,
with the gold-labeled cell highlighted in orange~(only during validation dataset).
Hovering the mouse over a cell reveals its cell ID near the cursor.
This interactive feedback allows the annotator to easily and accurately
pinpoint the cell ID
that may contain the target information throughout the annotation process.

\section{Results and Discussions}

\begin{table*}[t]
  \centering
  \caption{Results for test dataset in the TQA dataset.
  GPT-4o mini was utilized only for value extraction.
  }
  \begin{tabular}{lrrrrrr}
    \\ \hline
    model & T.C. &  U.E. &  V.N. & Cell ID & Value \\ \hline \hline
    GPT-4o mini  & $\times$  &  $\checkmark$ &  $\checkmark$ & -- & 40.6\% \\
    GPT-4o mini  & $\checkmark$  &  $\checkmark$ &  $\checkmark$ & -- & 63.9\% \\
    \hline
    TF-IDF & $\checkmark$ &  $\checkmark$ &  $\checkmark$ & 59.4\% & 58.7\% \\
    UBKE-LUKE & $\checkmark$ &  $\checkmark$ &  $\checkmark$ & 4.9\% & 6.5\% \\
    Sentence-BERT & $\checkmark$ &  $\checkmark$ &  $\checkmark$ & 31.2\% & 31.0\% \\
    Sentence-BERT~(FT) & $\checkmark$ &  $\checkmark$ &  $\checkmark$ & 75.9\% & 71.2\% \\
    TF-IDF~+~Sentence-BERT~(FT) & $\checkmark$ &  $\checkmark$ &  $\checkmark$ & \textbf{78.8\%} & \textbf{74.6\%} \\
    \hline
    Human & -- &  -- &  -- & 93.1\% & 85.0\%$^{\mathrm{*}}$ \\ 
    \hline
    \multicolumn{6}{p{10cm}}{$^{\mathrm{*}}$The value accuracy of human annotation is calculated by normalizing the human-extracted value with the unit extracted by U.E.}
  \end{tabular}
  \label{tab:test_result}
\end{table*}

The results of the TQA test dataset are shown in Table~\ref{tab:test_result}.

Vector retrieval without fine-tuning showed lower accuracy~(UBKE-LUKE: 6.5\% and Sentence-BERT: 31.0\% in value extraction)
than TF-IDF-based retrieval~(58.7\%).
Since a simple surface-level matching approach outperformed a pre-trained LM without domain-specific adaptation,
it can be said that, in the TQA dataset,
the ability to locate cells
containing words identical to those in the question is critical.

On the other hand,
GPT-4o mini,
which can handle a certain degree of structural understanding and semantic inference beyond mere surface matching,
achieved higher accuracy~(63.9\%)
than TF-IDF~(58.7\%).
However,
frequent errors were observed in which GPT-4o mini
chose the wrong row or column,
often mixing up fiscal years or item names.

In contrast,
the vector retrieval approach with fine-tuning through contrastive learning~(71.2\%)
outperformed GPT-4o mini (63.9\%).
This is likely because the model learned domain-specific positive and negative examples from the securities reports,
enabling more precise discrimination
among cells with similar surface forms.
This training was feasible
even with the relatively small dataset of 1,000 TQA questions
because we used a lightweight encoder model
(Sentence-BERT with 12 Transformer layers).

The hybrid search method,
which combines the fine-tuned Sentence-BERT with the TF-IDF retrieval,
outperformed all other approaches~(74.6\%).
We attribute this to the strategy of
uniting the fine-tuned sentence encoder with TF-IDF,
thereby leveraging both semantic and surface-level matching capabilities
to achieve stable cell identification and value extraction in the TQA dataset.

Furthermore,
simply cleaning HTML tables
before applying these approaches
significantly reduced noise and improved accuracy
across all methods.
For example, the value accuracy of the GPT-4o mini
increased by 23.3\% after the table cleaning process.
Our analysis also revealed that
while GPT-4o mini can extract numerical values or dates
in header form relatively well,
it often fails to identify the correct cell location in the table.

Our human-annotated evaluation
showed that
the non-expert in finance
achieved an accuracy of 93.1\% in extracting the required information from given tables.
The value accuracy of the human annotation was 85.0\%,
calculated by normalizing the extracted values using units determined by the U.E. step.

Even when compared to a human annotator without domain expertise,
all the methods examined in this study, i.e., both our methods and the LLM-based approaches,
still exhibit relatively low accuracy.
This finding suggests that these methods
continue to face challenges
in extracting information
that humans can intuitively understand
from structurally complex documents, such as securities reports.

\section{Conclusions and Future Work}
In this paper, we propose and evaluate
a pipeline to process complex tables in securities reports.
Unlike existing methods, our approach does not assume any prior identification
of table headers, even when the headers are highly complex.
It consists of:
(1) identifying the relevant rows and columns based on the given question,
and
(2) extracting the intersecting cell as the answer.
To achieve robust candidate retrieval,
we adopt a hybrid strategy that combines TF-IDF with
LM-based vector search.
In addition, we improve cell identification accuracy
by applying contrastive learning to a language model,
using domain-specific positive and negative examples.
As a result, on the TQA dataset,
our method achieved 78.8\% accuracy in cell ID extraction
and 74.6\% in value extraction,
surpassing the performance of large language models such as GPT-4o mini.

On the other hand,
a human annotator without specialized economic knowledge
achieved more than 93\% accuracy in information extraction.
Even when compared with a non-expert,
our methods and LLM-based approaches exhibit limited accuracy,
underscoring the challenge of extracting intuitively understandable information
from structurally complex documents.

For the future work,
although we used traditional encoder models for retrieval in this study,
we plan to incorporate more efficient text-search models
such as E5~\cite{e5} and Ruri~\cite{ruri}
to improve performance and narrow the gap with human evaluation results.
We will also conduct a detailed analysis of how humans versus automated systems
understand complex tables.

\section*{Acknowledgments}
We gratefully thank the U4 organizers in the NTCIR-18 conference for designing the task and providing the dataset that made our experiments possible. All results reported in this paper were obtained through experiments conducted on the dataset provided by the shared U4 task.

\newpage

\section*{Appendix A: Preliminary Experiment using GPT-4o mini} \label{app1}

We conducted a preliminary experiment to evaluate how accurately GPT-4o mini can answer questions
from the TQA dataset in U4 shared task~\cite{ntcir18-u4-overview},
using whole HTML-table data from securities reports.

Specifically, we provided GPT-4o mini with a given question
(e.g., ``What was the sales revenue in FY 2021?'')
and the corresponding HTML data including the table,
then measured how closely its generated response matched the correct cell value
for the TQA validation dataset.

As a result,
GPT-4o mini achieved an accuracy of approximately 32.1\% in extracting the correct value.
This finding suggests that GPT-4o mini
does have a certain capability to infer relevant items from table data.
However, a detailed examination of incorrect responses revealed two main types of errors:

\begin{itemize}
  \item Incorrect year:
  When multiple rows for the same item existed across different fiscal years, GPT-4o mini sometimes referenced the value from an unintended year.
  \item Incorrect value:
  When asked about a specific numerical figure (e.g., sales revenue, operating income), GPT-4o mini occasionally misinterpreted the target term and referred to another cell with a similar meaning.
\end{itemize}

In most cases,
these errors were caused by selecting the wrong row or column,
underscoring the importance of reliably identifying the correct row and column within the table.

The issues observed in this experiment motivated
the design of our proposed method,
which centers on estimating rows and columns to accurately extract the target cell.

\section*{Appendix B: Prompts for GPT-4o mini in U.E. and Cell ID Detection} \label{app2}
We input the following prompt into GPT-4o mini
in order to detect the correct cell ID for the question in the TQA dataset.
\begin{tcolorbox}
  Prompt =

  In the table below, please output the cell id of the cell that contains the answer to the following question.
  Please answer only the value of cell id.

  \# question: \{question\}

  \# table: \{table\}
\end{tcolorbox}

We input the following prompt into GPT-4o mini
in order to extract value from a table for the question in the TQA dataset.
\begin{tcolorbox}
  System Prompt =
  
  You are an AI assistant for detecting information.

  Prompt=

  In the table below, please answer the question.
  Please output the value and its unit with the format:
  
  \{``value'': \textless value\textgreater, ``unit'': \textless unit\textgreater\}

  \# question: \{question\}

  \# table: \{table\}
\end{tcolorbox}


\begin{thebibliography}{99}
  \bibitem{chen-etal-2021-finqa}
  Z. Chen, W. Chen, C. Smiley, S. Shah, I. Borova, D. Langdon, R. Moussa, M. Beane, T. H. Huang, B. Routledge and W. Y. Wang,
  ``FinQA: A Dataset of Numerical Reasoning over Financial Data,''
  in Proceedings of the 2021 Conference on Empirical Methods in Natural Language Processing, Online and Punta Cana,
  2021, pp. 3697--3711.
  \bibitem{koval-etal-2024-financial} R. Koval, N. Andrews and X. Yan,
  ``Financial Forecasting from Textual and Tabular Time Series,''
  in Proceedings of Findings of the Association for Computational Linguistics: EMNLP 2024,
  Miami, Florida, USA, 2024, pp. 8289--8300.
  \bibitem{wang-sun-2022-promptehr} Z. Wang and J. Sun,
  ``PromptEHR: Conditional Electronic Healthcare Records Generation with Prompt Learning,''
  in Proceedings of the 2022 Conference on Empirical Methods in Natural Language Processing, 2022, pp. 2873--2885.
  \bibitem{9856660} Y. Qu, F. Li, L. Li, X. Dou and H. Wang,
  ``Can We Predict Student Performance Based on Tabular and Textual Data?,''
  in IEEE Access, vol. 10, 2022, pp. 86008--86019.
  \bibitem{Lu_2024_tableLLM_survey} W. Lu, J. Zhang, J. Fan, Z. Fu, Y. Chen and X. Du,
  ``Large language model for table processing: a survey,'' 
  in Springer-Verlag, Berlin, Heidelberg, vol. 19, no. 2, 2025.
  \bibitem{fang2024large} X. Fang, W. Xu, F. A. Tan, Z. Hu, J. Zhang, Y. Qi, S. H. Sengamedu and C. Faloutsos,
  ``Large Language Models (LLMs) on Tabular Data: Prediction, Generation, and Understanding - A Survey,''
  in Transactions on Machine Learning Research, 2024.
  \bibitem{Li_TableGPT} P. Li, Y. He, D. Yashar, W. Cui, S. Ge, H. Zhang, D. R. Fainman, D. Zhang and S. Chaudhuri,
  ``Table-GPT: Table Fine-tuned GPT for Diverse Table Tasks,''
  in ACM Manag. Data, vol. 2, no. 3, 2024.
  \bibitem{chen_FewShotTableReasoners} W. Chen,
  ``Large Language Models are few(1)-shot Table Reasoners,''
  in Proceedings of Findings of the Association for Computational Linguistics: EACL 2023,
  Dubrovnik, Croatia, 2023, pp. 1120--1130.
  \bibitem{glass-etal-2021-capturing} M. Glass, M. Canim, A. Gliozzo, S. Chemmengath, V. Kumar, R. Chakravarti, A. Sil, F. Pan, S. Bharadwaj and N. R. Fauceglia,
  ``Capturing Row and Column Semantics in Transformer Based Question Answering over Tables,''
  in Proceedings of the 2021 Conference of the North American Chapter of the Association for Computational Linguistics: Human Language Technologies,
  Online, 2021, pp. 1212--1224.
  \bibitem{sui-etal-2024-tap4llm} Y. Sui, J. Zou, M. Zhou, X. He, L. Du, S. Han and D. Zhang,
  ``TAP4LLM: Table Provider on Sampling, Augmenting, and Packing Semi-structured Data for Large Language Model Reasoning,''
  in Proceedings of Findings of the Association for Computational Linguistics: EMNLP 2024,
  Miami, Florida, USA, 2024, pp. 10306--10323.
  \bibitem{Tay_survey} Y. Tay, M. Dehghani, D. Bahri and D. Metzler,
  ``Efficient Transformers: A Survey,''
  in ACM Comput. Surv., vol. 55, no. 6, New York, NY, USA, 2022.
  \bibitem{ntcir18-u4-overview} Y. Kimura, E. Sato, K. Kadowaki and H. Ototake,
  ``Overview of the NTCIR-18 U4 Task,''
  in Proceedings of the 18th NTCIR Conference on Evaluation of Information Access Technologies, 2025.
  \bibitem{li-etal-2025-graphotter} Q. Li, C. Huang, S. Li, Y. Xiang, D. Xiong and W. Lei,
  ``GraphOTTER: Evolving LLM-based Graph Reasoning for Complex Table Question Answering,''
  in Proceedings of the 31st International Conference on Computational Linguistics,
  Abu Dhabi, UAE, 2025, pp. 5486--5506.
  \bibitem{pang-etal-2024-uncovering} C. Pang, Y. Cao, C. Yang and P. Luo,
  ``Uncovering Limitations of Large Language Models in Information Seeking from Tables,''
  in Proceedings of Findings of the Association for Computational Linguistics: ACL 2024,
  Bangkok, Thailand, 2024, pp. 1388--1409.
  \bibitem{Fang2012} J. Fang, P. Mitra, Z. Tang and C. L. Giles,
  ``Table header detection and classification,''
  in Proceedings of the Twenty-Sixth AAAI Conference on Artificial Intelligence,
  Toronto, Ontario, Canada: AAAI Press, 2012, pp. 599--605.
  \bibitem{Nagy2016} G. Nagy and S. Seth,
  ``Table headers: An entrance to the data mine,''
  in 2016 23rd International Conference on Pattern Recognition (ICPR), 2016, pp. 4065--4070.
  \bibitem{jauhar-etal-2016-tables} S. K. Jauhar, P. Turney and E. Hovy,
  ``Tables as Semi-structured Knowledge for Question Answering,''
  in Proceedings of the 54th Annual Meeting of the Association for Computational Linguistics (Volume 1: Long Papers),
  Berlin, Germany, 2016, pp. 474--483.
  \bibitem{jin-etal-2023-enhancing} N. Jin, D. Li, J. Chen, J. Siebert and Q. Chen,
  ``Enhancing Open-Domain Table Question Answering via Syntax- and Structure-aware Dense Retrieval,''
  in Proceedings of the 13th International Joint Conference on Natural Language Processing and the 3rd Conference of the Asia-Pacific Chapter of the Association for Computational Linguistics (Volume 2: Short Papers),
  2023, pp. 157--165.
  
  \bibitem{sun2016table} H. Sun, H. Ma, X. He, S. W. T. Yih, Y. Su and X. Yan,
  ``Table Cell Search for Question Answering,''
  in Proceedings of the companion publication of the 25th international conference on World Wide Web,
  2016, ACM - Association for Computing Machinery.
  \bibitem{herzig-etal-2020-tapas} J. Herzig, P. K. Nowak, T. Müller, F. Piccinno and J. Eisenschlos,
  ``TaPas: Weakly Supervised Table Parsing via Pre-training,''
  in Proceedings of the 58th Annual Meeting of the Association for Computational Linguistics,
  Online, 2020, pp. 4320--4333.
  \bibitem{yin-etal-2020-tabert} P. Yin, G. Neubig, W. Yih and S. Riedel,
  ``TaBERT: Pretraining for Joint Understanding of Textual and Tabular Data,''
  in Proceedings of the 58th Annual Meeting of the Association for Computational Linguistics,
  Online, 2020, pp. 8413--8426.
  \bibitem{openai2024gpt4technicalreport} OpenAI, ``GPT-4 Technical Report,'' arXiv:2303.08774, 2024.
  \bibitem{Lu2024TableSurvey} W. Lu, J. Zhang, J. Zhang and Y. Chen,
  ``Large Language Model for Table Processing: A Survey,'' CoRR, 2024.
  \bibitem{wang2024chain} Z. Wang, H. Zhang, C. L. Li, J. M. Eisenschlos, V. Perot, Z. Wang, L. Miculicich, Y. Fujii, J. Shang, C. Y. Lee and T. Pfister,
  ``Chain-of-Table: Evolving Tables in the Reasoning Chain for Table Understanding,''
  in ICLR, 2024.
  \bibitem{ChainOfThought} J. Wei, X. Wang, D. Schuurmans, M. Bosma, B. Ichter, F. Xia, E. H. Chi, Q. V. Le and D. Zhou,
  ``Chain-of-thought prompting elicits reasoning in large language models,''
  in Proceedings of the 36th International Conference on Neural Information Processing Systems,
  New Orleans, LA, USA: Curran Associates Inc., 2022, pp. 1800.
  \bibitem{hurst2024gpt4o} A. Hurst, A. Lerer, A. P. Goucher, A. Perelman, A. Ramesh, A. Clark, A. J. Ostrow, A. Welihinda, A. Hayes and A. Radford,
  ``Gpt-4o system card,'' arXiv preprint arXiv:2410.21276, 2024.
  \bibitem{team2023gemini} Gemini Team, R. Anil, S. Borgeaud, J. B. Alayrac, J. Yu, R. Soricut, J. Schalkwyk, A. M. Dai, A. Hauth and K. Millican,
  ``Gemini: a family of highly capable multimodal models,'' arXiv preprint arXiv:2312.11805, 2023.
  \bibitem{Claude3} AI Anthropic, ``The Claude 3 Model Family: Opus, Sonnet, Haiku,''
  [Online]. Available: https://api.semanticscholar.org/CorpusID:268232499.

  \bibitem{reimers-gurevych-2019-sentence} N. Reimers and I. Gurevych,
  ``Sentence-BERT: Sentence Embeddings using Siamese BERT-Networks,''
  in Proceedings of the 2019 Conference on Empirical Methods in Natural Language Processing and the 9th International Joint Conference on Natural Language Processing (EMNLP-IJCNLP),
  Hong Kong, China: Association for Computational Linguistics, 2019, pp. 3982--3992.
  \bibitem{yamada-etal-2020-luke} I. Yamada, A. Asai, H. Shindo, H. Takeda and Y. Matsumoto,
  ``LUKE: Deep Contextualized Entity Representations with Entity-aware Self-attention,''
  in Proceedings of the 2020 Conference on Empirical Methods in Natural Language Processing (EMNLP),
  Online: Association for Computational Linguistics, 2020, pp. 6442--6454.
  \bibitem{Kudo2005MeCabY} T. Kudo, ``MeCab: Yet Another Part-of-Speech and Morphological Analyzer,'' 2005.
  \bibitem{kingma2017} D. P. Kingma and J. Ba, ``Adam: A Method for Stochastic Optimization,'' arXiv:1412.6980, 2017.
  \bibitem{e5} L. Wang, N. Yang, X. Huang, L. Yang, R. Majumder and F. Wei,
  ``Improving Text Embeddings with Large Language Models,''
  in Proceedings of the 62nd Annual Meeting of the Association for Computational Linguistics (Volume 1: Long Papers),
  2024, pp. 11897--11916.
  \bibitem{ruri} H. Tsukagoshi and R. Sasano,
  ``Ruri: Japanese General Text Embeddings,''	arXiv:2409.07737, 2024.
\end{thebibliography}
\end{document}